# Système de Reconnaissance Automatique de l'arabe basé sur CMUSphinx

H. Satori [1, 2], M. Harti [1, 2], and N. Chenfour [1, 2].

*(1) : UFR Informatique et Nouvelles Technologies d'Information et de Communication B.P. 1796, Dhar Mehraz Fès Morocco.*

*(2) : Département de Mathématiques et Informatique, Faculté des Sciences, B.P. 1796, Dhar Mehraz Fès, Morocco*
E-mail: hsnsatori@yahoo.fr

*Abstract*

*In this paper we present the creation of an Arabic version of Automated Speech Recognition System (ASR). This system is based on the open source Sphinx-4, from the Carnegie Mellon University. Which is a speech recognition system based on discrete hidden Markov models (HMMs). We investigate the changes that must be made to the model to adapt Arabic voice recognition.*

*Résume.*

*Dans ce travail nous allons réaliser un système de Reconnaissance Automatique de la Parole (RAP) basé sur le CMU Sphinx4. Ce dernier est un projet Open Source de l'Université Carnegie Mellon. Nous allons démontrer l'adaptabilité de ce système pour la reconnaissance de la langue arabe.*

**Mots-clés:** *Reconnaissance de la parole, Modèle acoustique, Langue arabe, HMMs, CMUSphinx-4, Intelligence artificielle.*

## 1. Introduction

La Reconnaissance Automatique de la Parole (RAP) est une technologie informatique permettant à un logiciel d'interpréter une langue naturelle humaine. Elle permet à une machine d'extraire le message oral contenu dans un signal de parole. Cette technologie utilise des méthodes informatiques des domaines du traitement du signal et de l'intelligence artificielle [1]. Les applications qu'ont peut imaginer sont nombreuse : aider les personnes handicapées, contrôle vocal des machines, réservation des vols, apprentissage d'autres langues, etc. [2].

Vue l'importance de la RAP, plusieurs systèmes ont été développés pour la reconnaissance vocale, parmi les plus connus: Dragon Naturally Speaking, IBM Via voice, Microsoft SAPI et d'autre. Aussi, il y a des Open Sources comme HTK [3], ISIP [4], AVCSR [5] et CMU Sphinx [6-8]. Nous sommes intéressés à ce dernier qui est un système basé sur les Modèles de Markov Cachés (MMC), en anglais Hiden Markov Models (HMM) [9]. Nous avons constaté que le système de reconnaissance de la parole CMU Sphinx 4 est librement disponible (Open Source) et il est actuellement l'un des systèmes de reconnaissance de parole les plus puissants. Le CMU Sphinx permet à des groupes de recherche avec des budgets modestes de développer et de conduire des applications de recherches dans la reconnaissance de la parole. Pour ces raisons et d'autres, nous avons choisi ce système pour développer notre application pour la reconnaissance de la langue arabe [10-11].

Notre travail s'inscrit dans le cadre général de la RAP il est parmi les premiers travaux traitant la langue arabe utilisant l'Open Source CMU Sphinx. Nous présentons dans ce travail les bases de construction d'un système de reconnaissance automatique de l'arabe classique basée sur le CMU Sphinx4.

## 2. Présentation du CMU Sphinx 4

Sphinx est un projet lancé par l'université Carnegie Mellon (CMU) dans le but de concevoir un environnement pour la recherche dans le domaine de la reconnaissance automatique de la parole. CMU Sphinx 4 est une librairie de classes et d'outils disponible en langage de programmation Java. Cette librairie est gratuite à télécharger, elle vice principalement à faciliter la construction des systèmes de reconnaissance vocale. CMU Sphinx-4 est un système de RAP basé sur les Models de Markov Cachés (HMM). Il a été créé conjointement par le groupe Sphinx à l'université CMU, les laboratoires Sun Microsystems et Hewlett-Packard company [12-14].

SphinxTrain est l'outil crée par CMU pour le développement des modèles acoustiques. C'est un ensemble de programmes et documentations pour réaliser et construire des modèles acoustiques pour n'importe quelle langue.

### 2.1. Architecture

Sphinx-4 présente un ensemble d'outils de reconnaissance vocale (voir figure 1) flexibles modulaires et extensibles formant un véritable banc d'essais et un puissant environnement de recherche pour les technologies de reconnaissance automatique de la parole.

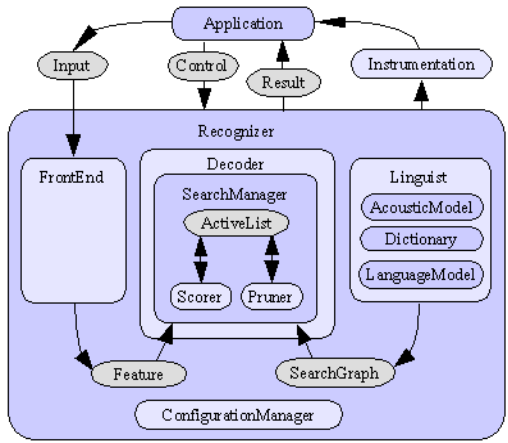

**Fig. 1:** Architecture du CMU Sphinx-4.

• **FrontEnd** : découpe la voix enregistrée en différentes parties et les prépare pour le décodeur. Il est responsable de la génération des vecteurs caractéristiques représentant les caractéristiques du signal vocal.
• **Features :** et utilisé pour estimer les paramètres du modèle acoustique.
• **Linguist :** ou base de connaissances qui est l'information qu'utilise le décodeur pour déterminer les mots et les phrases prononcées, elle est composée de :
– Dictionary.
– AcousticModel : modèle acoustique, un modèle statistique décrivant la distribution des données de phonèmes.
– LanguageModel : un modèle de langage, il donne la probabilité d'apparition d'un mot donné, basée sur des connaissances tirées du Dictionnaire.
• **SearchGraph** : contient toutes les séquences de phonèmes possibles basées sur le LanguageModel.
• **Decoder** : ou Décodeur qui est le coeur de Sphinx-4 ; c'est lui qui traite les informations reçues depuis le FrontEnd, il les analyse et les compare avec la base de connaissances pour donner un résultat à l'application.

## 2.2. Installation

### 2.2.1 Sphinx-4

Sphinx-4 peut être téléchargé de l'internent soit sous forme binaire soit sous forme source code [15]. Il a été compilé et testé sur plusieurs versions de Linux et sur Windows. L'exécution de Sphinx-4 demande des logiciels supplémentaires qui sont :

• Java 2 SDK, Standard Edition 5.0 [16].
• Java Runtime Environement (JRE)
• Les différentes librairies qui composent Sphinx-4.
• Ant : L'outil pour faciliter la compilation en automatisant les taches répétitives [17].

### 2.2.2 Sphinxtrain

SphinxTrain téléchargeable dont le lient se trouve dans tools du site de CMU Sphinx[13].
Les différentes librairies qui composent SphinxTrain :
• ActivePerl : L'outil pour éditer des scriptes pour SphinxTrain et permet de travailler dans un Unix-like environnement pour Windows plateforme [18].
• Microsoft Visual Studio : Pour compiler les sources en C afin de produire les Exécutables.

## 3. Reconnaissance de la langue arabe

La langue arabe est une langue sémitique, elle est parmi les langues les plus anciennes dans le monde [19].
L'arabe classique standard a 34 phonèmes parmi lesquels 6 sont voyelles et 28 sont des consonnes [20]. Les phonèmes arabe se distinguent par la présence de deux classes qui sont appelées pharyngales et emphatiques. Ces deux classes sont caractéristiques des langues sémitiques comme l'hébreu [20-22].
Les syllabes permises dans la langue arabe sont : CV, CVC et CVCC. Où le V désigne voyelle courte ou longue et le C représente une consonne [20].
La langue arabe comporte cinq types de syllabes classées selon les trais ouvert/fermé et court/long. Une syllabe est dite ouverte (respectivement fermée) si elle se termine par une voyelle (respectivement une consonne). Toutes les syllabes commencent par une consonne suivie d'une voyelle et elles comportent une seule voyelle. La syllabe CV peut se trouver au début, au milieu ou à la fin du mot [22-25].

### 3.1 Corpus

Le corpus est constitué des dix premiers chiffres de l'arabe classique de 0 à 9. Six locuteurs marocaines, 3 males et 3 femelles, sont invités à prononcer les dix chiffres cinq fois. Le corpus comprend cinq répétitions par chaque locuteur du même chiffre. Ainsi, le corpus est constitué de 300 tokens (10 chiffres. 5 répétitions. 6 locuteurs).
Pendant l'enregistrement, chaque répétition a été rejouée pour s'assurer que le chiffre entier a été inclus dans le signal enregistré. Dans le tableau 1 sont donnés certains paramètres d'enregistrement du corpus.

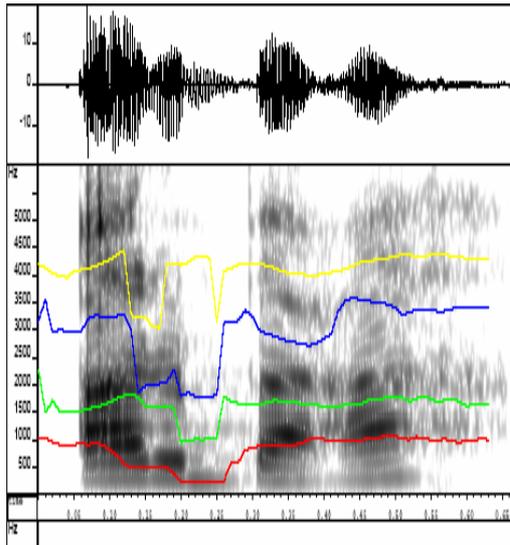

**Fig. 2:** Spectrogramme du chiffre 4 (أربعة) locuteur 2 essai 2, généré par l'open source wavesurfer [26].

| Paramètre | Valeur |
|---|---|
| Echantillonnage | 16 kHz, 16 bits |
| Wave format | Mono, wav |
| Corpus | 10 chiffres arabes |
| Locuteur | 6 (3 males + 3 femelles) |

**Tableau 1**: Paramètres d'enregistrement utilisés pour la préparation du corpus Arabic digits.

### 3.2 Modèle acoustique (Acoustic model)

Le modèle acoustique est une représentation statistique de l'image acoustique la plus significative possible pour le signal vocale. Durant la phase d'apprentissage, training, chaque unité acoustique ou phonème est représentée par un modèle statistique décrivant la distribution des données. Le signal parole est transformé en une série de vecteurs de caractéristiques (feature vectors) comprenant les coefficients MFCC (Mel-Frequency Cepstral Coefficients) [27].

| Symbole | Alphabet | Translittération |
|---|---|---|
| AA | ء | Alef |
| B | ب | Ba' |
| T | ت | Ta' |
| TH | ث | Tha' |
| HH | ح | Ha' |
| KH | خ | Emphatique Kha' |
| D | د | Dal |
| R | ر | Ra' |
| AIN | ع | Ayn |
| S | س | Sin |
| SS | ص | Emphatique Sad |
| L | ل | Lam |
| M | م | Mim |
| H | ه | Ha' |
| W | و | Waw |
| Y | ي | Ya' |
| A | َ | Fatha |
| I | ِ | Kasra |
| E | ِ | Entre kasra et fatha |

**Tableau 2:** Symboles de phonèmes, utilisé pour Arabic digits.

Dans notre application l'ensemble de symboles précèdent (voir tableau 2) a été utilisé pour l'apprentissage des états HMM correspondant au modèle acoustique de la démonstration Arabic_digits.

Le système doit savoir à quel HMM correspond chaque variable (phonème). Ces informations sont stockées dans un fichier appelé dictionnaire. Il permet de faire une représentation symbolique pour chaque mot. Il permet ainsi d'alimenter l'application Sphinxtrain pour produire le modèle acoustique. L'apprentissage a été faite en utilisant le dictionnaire représenté dans le tableau 3.

| 0 | SS E F R |
|---|---|
| 1 | W A A HH I D |
| 2 | AA I TH N A A N I |
| 3 | TH A L A A TH A H |
| 4 | AA A R B A A IN A H |
| 5 | KH A M S A H |
| 6 | S I T T A |
| 7 | S A B B AIN A |
| 8 | TH A M A A N I Y Y A |
| 9 | T I S AIN A |

**Tableau 3:** Extrait du fichier dictionnaire de l'application Arabic digits.

### 3.3 Modèle de langue (Language model)

Modèle de langue (Language model ou grammar model) c'est un modèle qui défini l'usage des mots dans une application. Chaque mot dans le modèle de lange doit être dans le dictionnaire de prononciation. Le choix d'un modèle de langue dépend de l'application, dans certains cas il n'est pas facile, ce n'est pas le cas dans notre démonstration arabic digits (voir fig. 3).

```
/**
 * JSGF Digits Grammar for Hello  Arabic Digits example
 */
grammar arabicdigits;

public <arabicdigits> (0 | 1 | 2 | 3 | 4 | 5 | 6 | 7 | 8 | 9)* ;
```

**Fig. 3** Extrait du fichier de grammaire de l'application Arabic digits.

### 3.4. Configuration du système Sphinx 4

Un système de reconnaissance automatique de la parole comme Sphinx 4 utilise deux éléments dépendant du langue: Le modèle acoustique et le modèle de langue.

Dans notre application nous avons procédé à la modification de ces deux modèles comme décrit précédemment.

Le sphinx 4 doit être configuré en utilisant un fichier xml. Ainsi le choix d'algorithmes, l'extraction et la comparaison de vecteurs de caractéristiques et d'autres aspects important pour la création d'un système de RAP peuvent être personnalisé au besoin et au choix de l'application considérée.

### 3.5 Résultas

Dans le but d'évaluer les performances de notre système, nous l'avons testé pour diffèrent locuteurs. Des personnes, des deux sexes, sont invitées à prononcer les dix chiffres arabes de 0 à 9. Nous avons enregistré le nombre de chiffres correctement reconnus, un taux moyen de reconnaissance a été calculé (voir tableau 4 et 5).

|    | Essai 1 | Essai 2 | Essai 3 | Taux de reconnaissance |
|----|---------|---------|---------|------------------------|
| M1 | 9       | 8       | 9       | 86,66%                 |
| M2 | 8       | 9       | 9       | 86,66%                 |
| M3 | 8       | 8       | 9       | 83,33%                 |
| W1 | 9       | 8       | 8       | 83,33%                 |
| W2 | 8       | 8       | 8       | 80,00%                 |
| W3 | 9       | 9       | 8       | 86,66%                 |

**Table 4:** Résultats du teste de l'application Arabic digits pour des locuteurs individuels, où M désigne Homme et W femme.

| Locuteurs | Nombre de locuteurs | Taux de reconnaissance |
|-----------|---------------------|------------------------|
| Male      | 3                   | 85,56%                 |
| Femelle   | 3                   | 83,34%                 |

**Table 5:** Taux de reconnaissance moyen pour des locuteurs des deux sexes.

Les résultats sont très satisfaisants vu la taille de notre corpus d'apprentissage qui est relativement petit. Il est recommandé de faire l'apprentissage (training) avec plus de 500 voix différentes [28] pour atteindre un taux de reconnaissance de 100%. Nous n'avons pas utilisé un corpus volumineux pour l'apprentissage, mais nos résultats sont déjà encourageants.

### 4. Conclusion

Pour conclure, un système de reconnaissance automatique de la langue a été conçu et adapté pour la reconnaissance de la langue arabe. Le système a été basé sur CMU Sphinx-4 de l'Université Carnegie Mellon. Une application, a été présentée pour démonter l'adaptabilité de ce système pour la langue arabe.

Dans les perspectives, nous projetons d'étendre l'application pour un large vocabulaire de la langue arabe. Aussi, la réalisation d'un système pour la reconnaissance du dialecte marocain.

### 5. Références